\documentclass{article}

\usepackage{PRIMEarxiv}

\usepackage[utf8]{inputenc} 
\usepackage{adjustbox}
\usepackage[T1]{fontenc}    
\usepackage{hyperref}       
\usepackage{url}            
\usepackage{booktabs}       
\usepackage{amsfonts}       
\usepackage{nicefrac}       
\usepackage{microtype}      
\usepackage{lipsum}
\usepackage{fancyhdr}       
\usepackage{graphicx}       
\graphicspath{{media/}}     
\usepackage{amsmath}
\usepackage{gensymb}
\usepackage{longtable}
\usepackage{subcaption}
\usepackage{pdfpages}
\usepackage{longtable}
\usepackage{setspace}
\usepackage{color}
\usepackage{cite} 

\title{Domain-informed operation excellence of gas turbine system with machine learning}

\author{
\begin{minipage}[t]{\textwidth}
\centering
\begin{tabular}{ccc}
\textbf{Waqar Muhammad Ashraf} & \textbf{Amir H. Keshavarzzadeh} & \textbf{Abdulelah S. Alshehri} \\
University College London & University of Cambridge & King Saud University \\
London, UK & Cambridge, UK & Riyadh, Saudi Arabia \\
The Alan Turing Institute & & \\
London, UK & & \\
\texttt{waqar.ashraf.21@ucl.ac.uk} & \texttt{ak2495@cam.ac.uk} & \texttt{aalshehri9@ksu.edu.sa} \\
\end{tabular}
\end{minipage}
\vspace{1.5em}
\\ 
\begin{minipage}[t]{\textwidth}
\centering
\begin{tabular}{ccc}
\textbf{Abdulrahman bin Jumah} & \textbf{Ramit Debnath} & \textbf{Vivek Dua}$^*$ \\
\textbf{King Saud University} & \textbf{University of Cambridge} & \textbf{University College London} \\
\textbf{Riyadh, Saudi Arabia} & \textbf{Cambridge, UK} & \textbf{London, UK} \\
\texttt{abinjumah@ksu.edu.sa} & \texttt{rd545@cam.ac.uk} & \texttt{v.dua@ucl.ac.uk} \\
\end{tabular}
\end{minipage}
}

\begin{document}
\maketitle

\begin{abstract}
The domain-consistent adoption of artificial intelligence (AI) remains low in thermal power plants due to the black-box nature of AI algorithms and low representation of domain knowledge in conventional data-centric analytics. In this paper, we develop a MAhalanobis Distance-based OPTimization (MAD-OPT) framework that incorporates the Mahalanobis distance-based constraint to introduce domain knowledge into data-centric analytics. The developed MAD-OPT framework is applied to maximize thermal efficiency and minimize turbine heat rate for a 395 MW capacity gas turbine system. We demonstrate that the MAD-OPT framework can estimate domain-informed optimal process conditions under different ambient conditions, and the optimal solutions are found to be robust as evaluated by Monte Carlo simulations. We also apply the MAD-OPT framework to estimate optimal process conditions beyond the design power generation limit of the gas turbine system, and have found comparable results with the actual data of the power plant. We demonstrate that implementing data-centric optimization analytics without incorporating domain-informed constraints may provide ineffective solutions that may not be implementable in the real operation of the gas turbine system. This research advances the integration of the data-driven domain knowledge into machine learning-powered analytics that enhances the domain-informed operation excellence and paves the way for safe AI adoption in thermal power systems.   
\end{abstract}


\section{Introduction}
Combined cycle gas power plants are crucial to allow for the shift of the energy mix from high to low emissions, especially in energy systems dominated by thermal plants \cite{davis2018net,bistline2022role}. Combined cycle gas power plants can rapidly increase their power output for peak energy demand, produce lower CO$_{2}$ emissions, and generally have higher energy efficiency than coal power plants \cite{wilson2018rapid}. Therefore, effective monitoring of gas power plant performance metrics, including thermal efficiency and turbine heat rate, is critical to maintaining an economic mode of operation that is complemented by a lower emissions discharge to the environment. Furthermore, accurate modeling and optimization of performance metrics for gas turbine systems is a fundamental task for effective operation of gas power plants that improves operational excellence and supports the achievement of the net zero target from the energy sector \cite{ieaai}.

In recent years, artificial intelligence (AI) and machine learning (ML) models have demonstrated good performance in the modeling of complex systems due to their ability and capacity to learn from hyperdimensional input space and big data \cite{deb2022machine,vlachas2022multiscale,qi2020using}. With reference to power systems and particularly to gas power plants, AI models, including artificial neural network (ANN), support vector machine, and ensemble methods, are deployed to model or forecast electric power generation and emissions discharge from combined cycle gas power plants \cite{dos2024co,saeed2023electrical,siddiqui2021power}. These studies are mainly based on open source data sets related to gas power plants available in the UC Irvine machine learning repository \cite{gas_turbine_co_and_nox_emission_data_set_551,combined_cycle_power_plant_294}.

Despite the excellent potential of AI to model complex functions, its adoption in real-time operation control and management for gas power plants remains marginally low \cite{ieaai}. The low adoption rate of AI in the safety critical operation of gas power plants stems from the inaccessibility of high-resolution industrial datasets of combined-cycle gas power plants, the black-box nature of most AI algorithms, the low explainability of model predictions, and insufficient integration of domain knowledge in data-centric analytics. A key barrier to AI adoption is also the lack of domain consistency in the results produced by AI-based optimization routines. When AI models are integrated into optimization frameworks without incorporating explicit domain-informed constraints, they can estimate optimized operating conditions, which can violate operating restrictions or physical laws \cite{ashraf2025domain}. The disconnect of domain knowledge from optimization problems undermines trust and limits the practical deployment of AI-based solutions in high-stakes industrial environments \cite{shobeiry2024ai}.  

Literature review is conducted to identify the state-of-the-art of applying AI/ML for modeling and optimization of the performance of the gas power plants. Research studies are reported for improving the predictive performance as well as enhancing the control of gas power plant's operation. Osegi et al. \cite{osegi2023evolutionary} developed an evolutionary programming and continual learning neural network approach to predict the power output from an open-cycle gas power plant. The effect of ambient temperature on the power generation from the gas turbine system was also investigated. Wu et al. \cite{wu2023prediction} trained ML models to predict CO-NO$_{x}$ emissions from a natural gas power plant. The proposed methodology built on the feature optimization scheme improved the predictive performance of the ML model by 18 \%– 67 \%. In another study, Manatura et al. \cite{manatura2025gas} utilised process variables and ambient conditions to train an ANN model for predicting the turbine heat rate of a gas turbine system. 

Researchers have also implemented empirical and fuzzy logic-based constraints for the power generation operation of gas power plants. Al-Shami et al. \cite{designs7040085}, Mamlook et al. \cite{mamlook2011fuzzy} and Bertini et al. \cite{bertini2010fuzzy} implemented fuzzy rule-based domain knowledge of the gas turbine system to guide optimization solvers to avoid unsafe and inefficient operating conditions for the power plant. Model predictive control was applied to improve the operation and efficiency of the power generation of the gas power plant \cite{mohamed2016predictive,hou2018novel,hou2020fuzzy}. These studies have reported the effectiveness of fuzzy logic, model predictive control, and empiric bounds on the process variables for the control of gas power plant. However, the constraints can be univariate, require rule-based manual tuning, and can neglect the multivariate interaction that is a typical operating characteristic of the gas power plant's operation. 

The literature review reveals that limited research is reported on embedding the trained AI/ML models in the optimization frameworks for estimating the optimal process conditions for the gas power plant's operation. Thus, a key challenge of domain representation in the optimization problems for the operation excellence of gas power plant remains largely unexplored. Since ML models are trained to predict the output variable(s) within the operating bounds, their predictive accuracy outside the training space can be low and uncertain \cite{ratz2024identifying}, making them less trustworthy for capacity enhancement of gas power plants. Moreover, robustness evaluation of the ML model-based predictions is a key requirement for industrial applications \cite{braiek2025machine,freiesleben2023beyond}, which is rarely reported for ML-based studies, analysing the performance of gas power plants.   

To overcome the issues of representing domain knowledge in the AI-based analysis framework, we develop a Mahalanobis distance-based constraint that quantifies domain knowledge of gas power plant's operation, and is embedded in the optimization routine. The joint distributions of the variables are learnt from the historical operation data and serve as domain knowledge of the system since the variation in the process conditions is governed by the physical laws, and the joint data distributions capture this information. Embedding the variable's dependency structure through the Mahalanobis distance as a constraint in the optimization problem overcomes the issues with manually tuned fuzzy rules and streamlines the introduction of domain knowledge into the optimization routines. The Mahalanobis distance-based constraint constructs a multivariate ellipsoid that dynamically reflects the internal correlation structures of the process variables. The satisfaction of the Mahalanobis constraints respects the multivariate interaction and dependencies, provides domain-compliant solutions, and reduces the risks of operational infeasibility of the estimated solutions. Without the integration of the Mahalanobis constraint in optimization problems, the black-box nature of AI models and the greedy nature of optimization solvers to reach the set objective of the formulated optimization problem may provide solutions that do not follow the variables dependencies and thus are ineffective. This is a potential barrier to effectively using the non-linear learning capabilities of AI models in the optimization environment and is rarely investigated in the literature to address this issue.

This study introduces the Mahalanobis Distance-based optimization (MAD-OPT) framework, a novel methodology that enforces alignment between the functional representation within ANN and the explicit variable dependency structure of physical systems through a Mahalanobis constraint. This framework is empirically validated on a 395 MW gas turbine system, demonstrating its capability to optimize thermal efficiency and turbine heat rate across the entire power ramp-up operating regime. A critical comparison reveals that omitting the Mahalanobis constraint results in domain-inconsistent solutions, underscoring its necessity for preserving physical plausibility. Furthermore, the MAD-OPT framework exhibits extrapolative performance, enabling ANN-based predictions beyond the turbine’s design capacity. The robustness of optimized solutions is rigorously evaluated via Monte Carlo simulations, while the broader implications for operational excellence and AI integration in gas power plant management are discussed, highlighting the framework’s potential to bridge data-driven modeling with industrial process optimization.

The rest of the paper is organized as follows: section~\ref{sec:method} describes the key stages included in the MAD-OPT framework. Results and the associated discussions related to ANN model development, data-centric optimization of gas turbine system, ramping up gas turbine system under different ambient conditions, robustness evaluation of optimal solutions, operating the gas turbine system beyond the capacity discharge limit and the subsequent robustness evaluation are presented in section~\ref{sec:results}. Later, conclusions, limitations and future work are explained in the paper.

\section{Data and Methods} \label{sec:method}

\subsection{Data acquisition}

The MAD-OPT framework is applied for data-centric operation excellence analysis of a 395 MW capacity gas turbine system. A brief description of the power generation operation of the gas turbine system is provided.

The airflow rate is maintained by the air compressor while the natural gas compressor passes the natural gas through the heaters to raise its temperature, and natural gas is injected into the combustion chamber. The hot flue gases are produced in the combustion chamber from the combustion of natural gas and are passed through the gas turbine, where they expand and drive the rotor of the generator for producing electrical power. The exhaust of the flue gases from the gas turbines is made to pass through a heat recovery steam generator (HRSG) in a combined cycle gas power plant where heat is recovered from the exhaust flue gases and a steam cycle is in operation for power production. The combined cycle gas power plants typically have higher energy efficiency than the open-cycle gas power plaplants,ere exhaust flue gases are discharged to the environment. 

We have conducted a literature review \cite{goyal2020prediction,saleel2021forecasting,tufekci2014prediction,liu2020gas} and have also taken feedback from process and performance engineers to identify the operationally relevant input variables of the process. We identify the relevant process input variables and the associated dataset from \cite{ashraf2024driving} in line with the suggestions of the plant's performance and process engineers. We have also augmented the data with another process variable, namely "Flue Gas Temperature at HRSG Inlet" to further enrich the dataset. The compiled dataset is utilised to model the performance variables of the gas turbine system, including power (MW), thermal efficiency (\%) and turbine heat rate (kJ / kWh)). Thermal efficiency is a ratio of electrical energy produced from the chemical energy of the gas. The turbine heat rate measures the amount of thermal energy required to produce one unit of electricity. The two performance variables have a direct link with the cost of operation of the power generation and are critically controlled to maintain efficient power generation and reduce emissions discharge from the power plant.  

During the power generation operation from the gas power plant, the process variables have very different operating ranges. A significant deviation between the operating scales of the process variables negatively affects the functional mapping between the input-output variables that are constructed in the data-driven or AI models. To address this issue, we have scaled the data associated with the process variables and the performance variables on the scale of [0,1] using the following equation:
\begin{equation}
X_i^{\text{scaled}} = \frac{X_i - (X_i)_{min}}{(X_i)_{max} - (X_i)_{min}} 
\end{equation}

here, $X_i^{\text{scaled}}$ is the variable that is transformed on a scale [0,1], and ($X_i)_{min}$ and ($X_i)_{max}$ represent the minimum and maximum values of the input variables, $X_i$, respectively. The operation of a gas power plant undergoes a smooth transition when power is increased or decreased. It is natural to have collinear process variables which are monitored during the power generation operation of a gas turbine system. A pairwise collinearity measurement allows exploration of the dependencies of the variables in a structured way, and all possible combinations of bivariate collinearity can be investigated. Pearson's correlation coefficient ($PCC$) measures the linear dependencies between the pair of variables, and correlation couplings between the process variables can be visualized using a heat map. The mathematical expression of $PCC$ is given as:

\begin{equation}
PCC_{xy} = \frac{\sum_{i=1}^N (x_i - \bar{x})(y_i - \bar{y})}{\sqrt{\sum_{i=1}^N (x_i - \bar{x})^2 \sum_{i=1}^N (y_i - \bar{y})^2}} 
\end{equation}

here, $x_i$ \& $y_i$ is a variable pair that has observations, $i = 1,2,3,\dots,N$. $\bar{x}_i$ and $\bar{y}_i$ are calculated as the mean of $x_i$ and $y_i$, respectively. A value of $PCC$ that reaches +1 is an indicator of strong positive correlation and vice versa. Thus, the nature of the correlation (positive or negative) can be identified by $PCC$. It is important to mention here that the calculated $PCC$ is not just an indication of the correlation structures that exist in the data set. Rather, the carefully chosen process-related variables provide the rationale for treating these correlation values as relevant information regarding the operating characteristics of the system. Thus, $PCC$ or relevant statistical metrics compute the data-centric quantification of domain knowledge present in the raw measured data. The quantification of data-driven domain knowledge as a statistical metric serves as a ground truth to evaluate the efficacy of data-driven optimization solutions, as explained in the results section.

\subsection{ANN Model Development and Feature Importance Analysis}

ANN is a universal function approximator \cite{csaji2001approximation} and can find nonlinear and hidden variable structures in data \cite{goswami2024fault,kumar2025learning,ansar2025comparison,kumar2023variance}. A key advantage of using the ANN model is the computational overhead memory requirements, which are reasonable to model the complex function space on ill-defined problems \cite{rumelhart1986learning}. A three-layer shallow ANN structure has been found useful for modeling various types of problem \cite{gueddar2012novel,jamil2024machine}, and it has been proved in the literature that a shallow ANN model can approximate any nonlinear function with arbitrary accuracy given that a sufficient number of neurons are embedded in the hidden layer \cite{haykin2009neural}. The error backpropagation is the working mechanism for tuning the parameters of ANN, and various optimization solvers can be deployed for this task. In this study, we have implemented the Adaptive Moment Estimation (ADAM) solver along with the weight decaying parameter ($\lambda_2$) \cite{kingma2014adam} which is treated as a hyperparameter. Furthermore, we have also implemented the regularization $L1$ in the loss function ($\lambda_1$ parameter) to avoid overfitting issues and to enhance the generalization capability of the ANN model. A data split ratio of 0.8 and 0.2 is used for data partitioning into training and testing datasets, respectively. We have implemented ANN training in Pytorch.

The trained ANN models are evaluated by the coefficient of determination ($R^2$) and the root mean squared error ($RMSE$) \cite{muhammad2020optimization} which are computed as:

\begin{equation}
R^2 = 1 - \frac{\sum_{i=1}^N (y_i - \hat{y}_i)^2}{\sum_{i=1}^N (y_i - \bar{y})^2}
\end{equation}
\begin{equation}
RMSE = \sqrt\frac{\sum_{i=1}^N (y_i - \bar{y_i})^2}{N}
\end{equation}

Here, $y_i$, $\hat{y}_i$ are the actual and model-based simulated responses, respectively; while, $\bar{y}_i$ is calculated as the mean of $y_i$. $R^2$ varies between 0 to 1 and is considered as accuracy measure. Whereas, $RMSE$ computes the mean error induced in the point-prediction made by the trained model. The smaller the $RMSE$ is, a well-trained model is developed and vice versa.

The feature importance analysis is performed using the Shapley Additive Explanations (SHAP) method. SHAP values are calculated using a game theory technique and compute the marginalized contribution of each feature of the model to the predictions \cite{lundberg2017unified}. SHAP is found to be useful in explaining the predictions made by AI models in various studies \cite{ansar2025comparison,tariq2025explainable} and is applied to establish the importance of the feature in predicting performance variables (power, thermal efficiency, and turbine heat rate) of the gas turbine system.  

\subsection{Mahalanobis distance as a Domain-Informed Constraint}
The Mahalanobis distance is a multivariate distance metric that measures how far a point lies from the mean of the distribution, following the correlation structure in the data \cite{mclachlan1999mahalanobis}. The Mahalanobis distance can be computed on a multivariate vector. Mathematically, the Mahalanobis distance is written as:

\begin{equation}
    d_{M}(\mathbf{x}) = \sqrt{(\mathbf{x} - \boldsymbol{\mu})^\top \Sigma^{-1} (\mathbf{x} - \boldsymbol{\mu})}
\end{equation}

here, $\mathbf{x}$ is a vector of variables that is made mean-centered through $(\mathbf{x} - \boldsymbol{\mu})$. $\Sigma$ is a covariance matrix and encodes how the variables vary together. The diagonal terms are the variances of the variables, while off-diagonal terms represent the correlations between the variables. Taking the inverse of $\Sigma$ scales the variables in each direction and removes the correlation between the variables. The mathematical operations compute the Mahalanobis distance ($d_{M}(\mathbf{x})$) and the large value of $d_{M}(\mathbf{x})$ means that the point is far from the mean $\mu$ and vice versa. 

It is important to note here that the highly correlated variables do not contribute much to the computation of $d_{M}(\mathbf{x})$ than the orthogonal change (less correlated variables). This preserves the elliptical shape of the data cloud, where a large variance along the ellipse is the highly correlated direction. Thus, the Mahalanobis distance makes an ellipsoid around the multivariate data distribution and is used for outlier detection \cite{ghorbani2019mahalanobis} or classification tasks \cite{gallego2013mahalanobis} when the observed point does not fall within the ellipsoid. 
$d_{M}(\mathbf{x})$ can be set as a constraint since the calculated distance indicates how many standard deviations the point is from the mean. For example, $d_{M}(\mathbf{x}) = 1$ means that the point $\mathbf{x}$ is one standard deviation away from $\boldsymbol{\mu}$. Algebraically, an ellipsoid is defined in the $p$-dimensional space, and the region of the ellipsoid is controlled by $\tau$ as expressed in the following equation. 
\begin{equation}
   (\mathbf{x}-\mu)^\top\Sigma^{-1}(\mathbf{x}-\mu)\le \tau^2
\end{equation}
Geometrically, the contour $d_M(\mathbf{x})=\tau$ is an ellipsoid centered on $\mu$ whose axes align with the principal components of $\Sigma$. Since $\tau$ adjusts the region of the ellipsoid, it is a natural setting to introduce the Mahalanobis distance as a constraint in the optimization problems for two reasons: (i) the Mahalanobis distance handles the multivariate dependencies and preserves the correlation structures. This is aligned with the operation characteristics of the gas power plant in which we have correlated variable dependencies, and we want to represent them in an optimization problem for estimating domain-informed solutions, and (ii) we can adjust the ellipsoid region and may extrapolate the design space of the power plant. For example, if a gas power plant has a designed power generation capacity of 380 MW, we may use multivariate information and $\tau$ to estimate the new process conditions to produce power beyond 380 MW.  

\subsection{Formulation of Optimization Problem and Robustness Evaluation of the Optimal Solutions}
Thermal efficiency and turbine heat rate are the two critical performance variables measured in the sustained state of power generation from the gas turbine system. The objective of the optimization problem is to maximize thermal efficiency and minimize turbine heat rate at the set value of power. Since thermal efficiency and turbine heat rate exhibit nonlinear operating characteristics with process variables, the optimization problem is formulated with a nonlinear programming framework. The optimization problem embeds the trained ANN models for power, thermal efficiency, and turbine heat rate. Moreover, the Mahalanobis distance-based constraint is also incorporated in the optimization problem which is written in the following.

Objective function: 
\[
\min_{\mathbf{x}} f(\mathbf{x}) = - f_{\text{TE}}(\mathbf{x}) + f_{\text{THR}}(\mathbf{x})
\]
Subject to
\[
h(\mathbf{x}) = 0
\]
\[
 (f_{\text{Power}}(\mathbf{x}) - Power_{\text{ Set Point}})^2 < \epsilon
\]
\[
(\mathbf{x}-\mu)^\top\Sigma^{-1}(\mathbf{x}-\mu)\le \tau^2
\]
\[
\mathbf{x} = \{x_1, x_2, \dots, x_m\} \tag{5}
\]
\[
\mathbf{x} \in \mathbf{X} \subseteq \mathbf{R}^n
\]
\[
\mathbf{x}^L \leq \mathbf{x} \leq \mathbf{x}^U
\]

here, $h(\mathbf{x})$ is the equality constraint, representing trained ANN models for power, thermal efficiency, and turbine heat rate. The trained ANN model predicts the responses on the scale of [0,1], so scaling is not performed for the terms included in the objective function. $Power_{\text{ Set Point}}$ is set by the user from the operating space of the gas turbine operation so that its square difference from $f_{\text{Power}}$ is minimized to $\epsilon$. The Mahalanobis distance-based constraint is also incorporated into the optimization problem to satisfy the variable dependencies while estimating an optimal solution for the objective function. $\mathbf{x}$ is a set of operating variables, $x_1$, $x_2$,\dots, $x_m$ that are continuous and serve as the dimensions of the search space. The lower bounds ($\mathbf{x}^L$) and the upper bounds ($\mathbf{x}^U$) are set in $\mathbf{x}$ to limit the search space for the solver and estimate a solution within the operating bounds. 

The optimization problem is analyzed by two approaches: (1) MAD-OPT framework and (2) when we did not embed the Mahalanobis constraint into the optimization problem. The two optimization problems are implemented in Pyomo and are solved by the Sparse Nonlinear OPTimizer (SNOPT) solver \cite{gill2005snopt} through GAMS. The same initial guesses are used to solve the two optimization problems, and then a systematic comparison is made to analyze the implementability of the estimated solution in the gas turbine system. 

The robustness of the estimated optimal solutions of the three performance variables is evaluated under perturbed process conditions. 1000 Gaussian noise observations are generated on 1 \% of the standard deviation of process variables and are added with the optimal process input conditions. This procedure is repeated for 50 rounds of Monte Carlo simulations to fully explore the effect of noise generation and its impact on the variability of the optimal values of the turbine heat rate and thermal efficiency with respect to the set power value. A 95 \% confidence interval is calculated (97.5$^{th}$ quantile - 2.5$^{th}$ quantile) for each round of Monte Carlo simulations to account for variation in the performance variables. A robust solution has a narrow confidence interval when evaluated on the perturbed process variables and vice versa.

\subsection{Extrapolating the ANN model}
The Mahalanobis distance measures the multivariate dependencies and handles the correlations that exist in the data. We have developed the Mahalanobis-based constraint in the optimization problem as an extrapolation framework to extrapolate the ANN. This is because when the optimization solver estimates the optimal process conditions in the extrapolated zone of ANN, the Mahalanobis distance-based constraint guides the solver to comply with variable dependencies and interactions that are established amongst the process variables. For this investigation, we took a subsample space from the actual data of the gas power plant, that is, actual gas turbine system data have power generation up to 395 MW and we separate the actual data up to 380 MW. The remaining data set, having power ranging from 381 MW to 395 MW, is separated to serve as the ground truth during extrapolation analysis. The subsample space is deployed to train the ANN models, which can predict power with reasonable accuracy up to 380 MW since the subsample space was exposed to the ANN models during the training and testing phases. Later, the extrapolation analysis is carried out on the set values of power of 385 MW, 390 MW, and 395 MW. 

\subsection{Verification of the MAD-OPT Framework}
The proposed MAD-OPT framework is applied for estimating the optimal process conditions during the power ramp-up and capacity enhancement of a 395 MW capacity gas turbine system. Contour plots are made of the process variables, and ellipses are drawn around the data distribution, which establish the feasible operating regions for the power plant's operation. The estimated optimal solutions are mapped on the contour plots in order to verify the efficacy of the optimal solutions for their implementability in the plant's operation. Moreover, the extrapolated solutions are also compared with the actual data of the power plant to verify their effectiveness and, in turn, the ability of the MAD-OPT framework for the power generation capacity enhancement of the gas turbine system.  

\section{Results \& Discussions} \label{sec:results}

\subsection{Gas Turbine System Operation}

Gas turbine systems are installed to meet the peak energy demand of the grid. Although gas power systems maintain power generation capacity discharge into the grid, thermal efficiency and turbine heat rate are the key performance metrics that affect the cost of operation of the gas turbine system. The two performance metrics are driven by the operating levels of the process variables. In this paper, we analyze three performance variables, namely, the generation of power from a gas turbine system (Power - MW), the thermal efficiency (TE -\%) and turbine heat rate (THR - kJ / kWh) of a gas turbine system of 395 MW capacity. The input variables of the process taken in this paper to analyze three performance variables of the gas turbine system are: Compressor Discharge Pressure (CDP - Psi), Gas Fuel Flow Rate (GFFR - lb/s), Fuel Gas Temperature at the inlet of the combustion chamber (FGT - $^{\circ}$F), Ambient Temperature ($^{\circ}$C), Ambient Pressure (AP - hPa), Ambient Humidity (AH -\%), Performance Heater Gas Outlet Temperature (PHGOT - $^{\circ}$F), Compressor Discharge Temperature (CDT - $^{\circ}$F), and Flue Gas Temperature at HRSG inlet (FGEXT - $^{\circ}$C). The descriptive statistics of the collected dataset like minimum, mean, maximum and standard deviation, are computed and are mentioned in Table \ref{table:1}.

\begin{table}[htbp]
\centering
\caption{Descriptive Statistics of Variables of Gas Turbine System}
\label{table:1}
\begin{adjustbox}{max width=\textwidth}
\begin{tabular}{llrrrr}
\toprule
\textbf{Variable} & \textbf{Unit} & \textbf{Minimum} & \textbf{Mean} & \textbf{Maximum} & \textbf{Standard Deviation} \\
\midrule
Compressor Discharge Pressure (CDP) & Psi       & 186   & 248   & 312   & 36.82 \\
Gas Fuel Flow Rate (GFFR)           & lb/s      & 29    & 39    & 50    & 5.65 \\
Fuel Gas Temperature (FGT)          & $^\circ$F & 484   & 513   & 535   & 14.93 \\
Ambient Temperature (AT)            & $^\circ$C & 20    & 26    & 34    & 3.67 \\
Ambient Pressure (AP)               & hPa       & 983   & 988   & 992   & 1.99 \\
Ambient Humidity (AH)               & \%        & 34    & 66    & 98    & 14.16 \\
Performance Heater Gas Outlet Temperature (PHGOT) & $^\circ$F & 400   & 411   & 425   & 2.47 \\
Compressor Discharge Temperature (CDT)    & $^\circ$F & 813   & 861   & 926   & 34.26 \\
Flue Gas Temperature at HRSG Inlet (FGEXT) & $^\circ$C & 629   & 659   & 673   & 15.90 \\
Thermal Efficiency (TE)             & \%        & 32.69    & 38.99    & 42.97    & 2.36 \\
Power Generation (Power)            & MW        & 185   & 297   & 395   & 59.32 \\
Turbine Heat Rate (THR)             & kJ/kWh    & 8377  & 9267  & 11022 & 579.46 \\
\bottomrule
\end{tabular}
\end{adjustbox}
\end{table}

\begin{figure}[htp]
    \centering
    \includegraphics[width=1.0\linewidth]{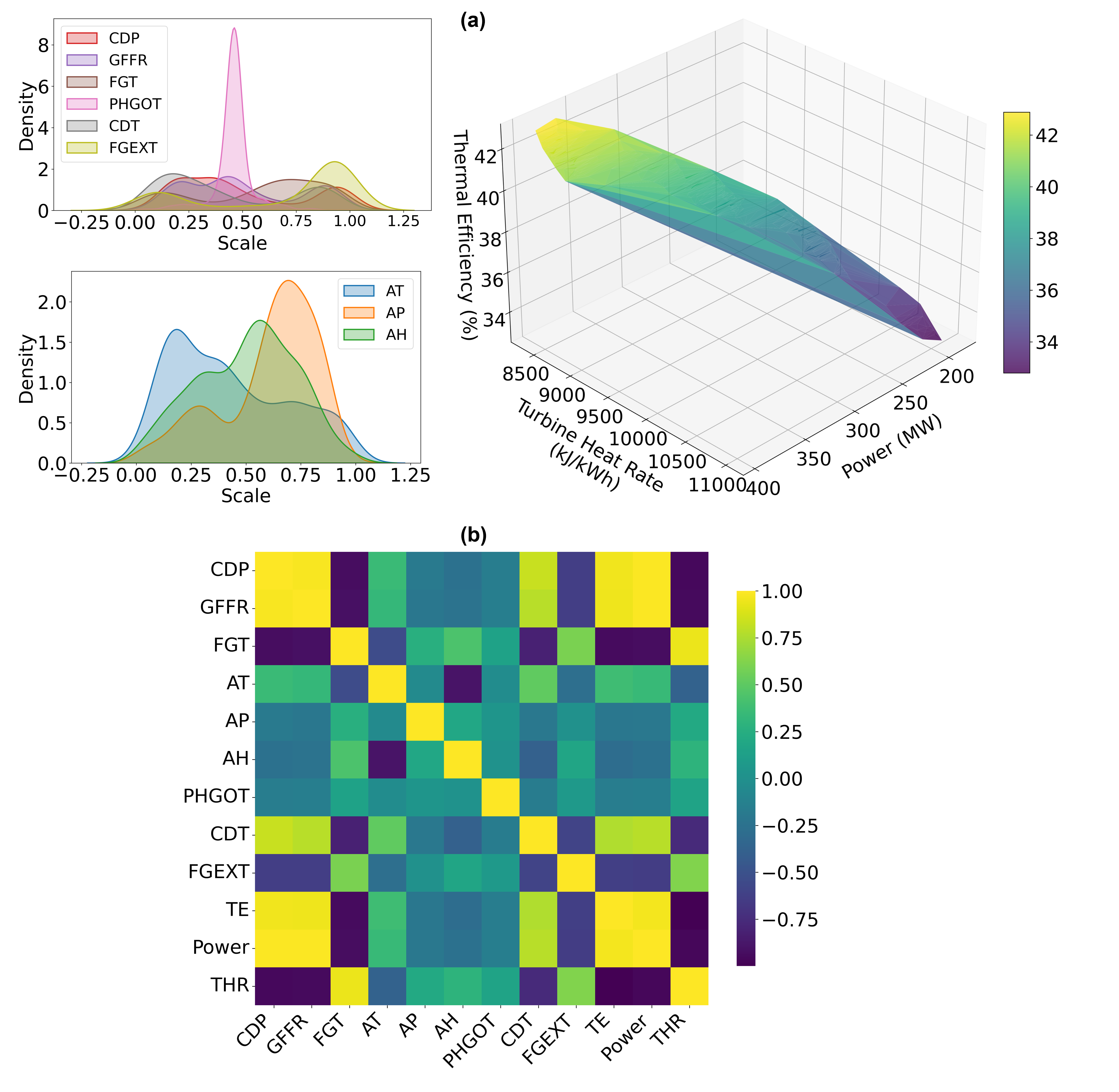}
    \caption{Power generation from a gas turbine system. (a) Data visualisation and (b) understanding the operation characteristics of system.}
    \label{Fig:Data_Visual}
\end{figure}

 The kernel density estimate (KDE) curves are plotted separately for ambient conditions (AT, AP, and AH) and process input variables (CDP, GFFR, FGT, PHGOT, CDT, and FGEXT) as shown in Fig.~\ref{Fig:Data_Visual}(a). The data density curves of the ambient conditions appear to be different from each other. However, process input variables seem to have nearly similar KDE curves except PHGOT. This is further confirmed with a heat map based on Pearson's correlation coefficients made on the pairs of process input and output variables, and is represented on Fig.~\ref{Fig:Data_Visual}(b). This demonstrates that process input variables have multivariate dependencies and correlation structures that characterize the power generation dynamics of the gas turbine system. In addition, the three performance variables are also correlated, depicting the operating behavior of the gas turbine system. To further explore the dependencies amongst the three performance variables, a function space is constructed which is found to be nonconvex and complex. Similarly, multivariate dependencies of varying degrees of correlation are observed with the three performance variables of the gas turbine system. This highlights the need to carry out robust data-driven analytics for the operation excellence of the gas turbine system.   

\subsection{Data-driven modeling of the Performance Variables of the Gas Turbine System}

\begin{figure}[htp]
    \centering
    \includegraphics[width=0.9\linewidth]{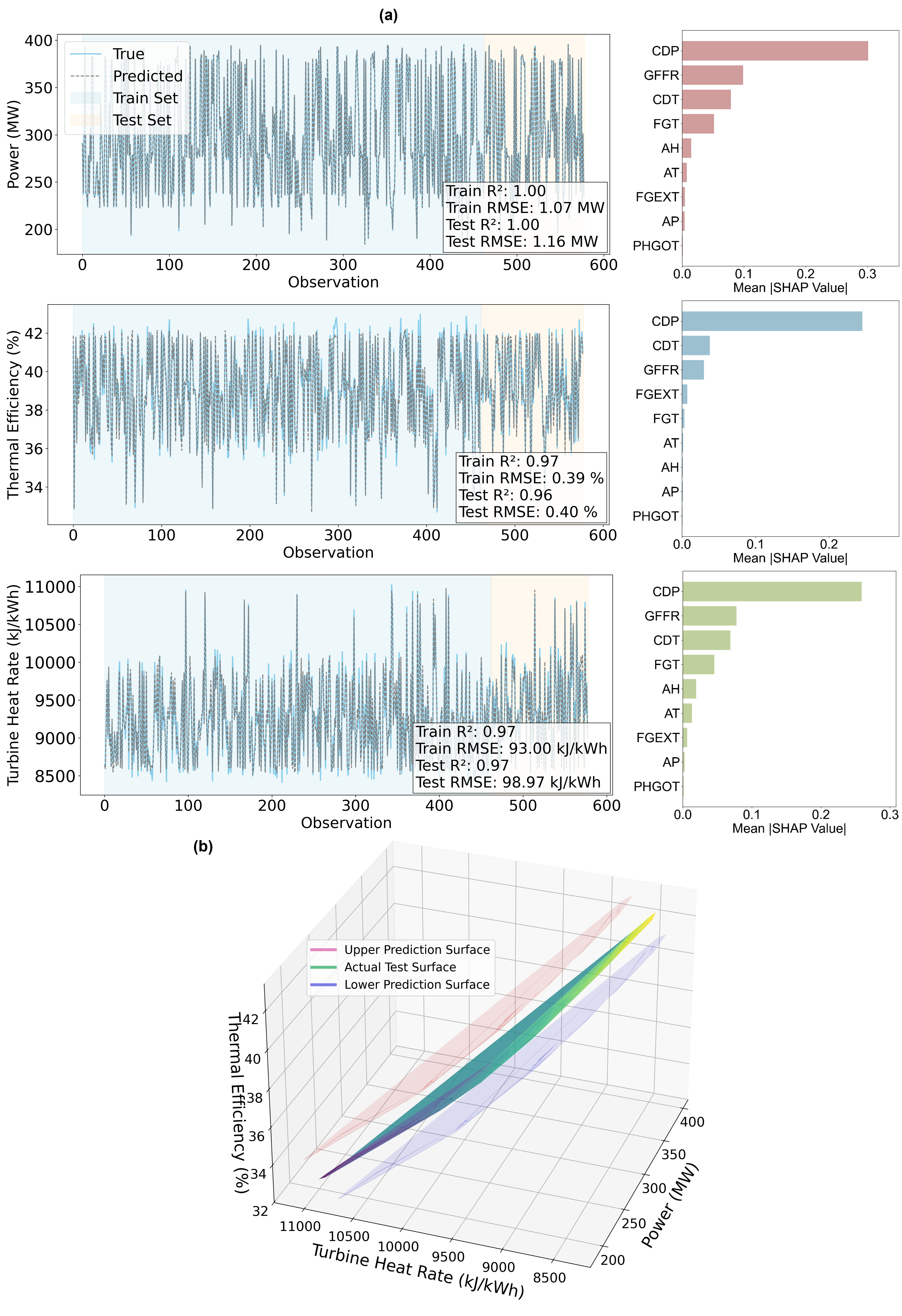}
    \caption{Data-driven modeling of the power generation system. (a) ANN-based modeling and the associated feature importance list made by SHAP for three key performance variables (thermal efficiency, power and turbine heat rate) of the gas turbine system. (b) Mapping the test data-driven actual function space of three performance variables on estimated upper and lower prediction surfaces.}
    \label{Fig:models}
\end{figure}

Fig.~\ref{Fig:models}(a) shows the modeling performance of ANN models trained to predict the thermal efficiency (\%), power (MW) and turbine heat rate (kJ / kWh) of gas turbine system. The trained ANN model for predicting thermal efficiency has 16 neurons in the hidden layer. The model predicts thermal efficiency with R$^{2}$ of 0.97 and 0.96, while RMSE of 0. 39 \% and 0. 40\% are obtained on the training and testing datasets, respectively. A perfect fit is observed on the training and testing data sets to predict the power (MW) using the trained ANN model (optimized hidden layer neurons are 31), measuring R$^{2}$ of 1.0, respectively. 

However, RMSE of 1.07 and 1.16 MW, respectively, are present in the predictions made by the power model in the training and testing datasets. The ANN model trained to predict the turbine heat rate has an optimal number of neurons in the hidden layer of 31, and the model predicts the turbine heat rate with R$^{2}$ of 0.97 and RMSE of 93 kJ/kWh and 99 kJ/kWh, respectively, on training and testing data sets. The trained ANN models to predict the three performance variables exhibit excellent predictive performance on test datasets, which is not only comparable with those of training datasets but is also the indicator of good generalisation capability induced in the trained models for the predictive tasks.

The feature importance analysis is performed by embedding the trained ANN models in the SHAP framework, and the feature importance order is established for the three performance variables. The mean SHAP values are plotted as bars corresponding to each variable and are shown on the right side of Fig.~\ref{Fig:models}(a). SHAP-based analysis reveals that CDP is the most significant process variable and is followed by GFFR, which affects the thermal efficiency of a gas turbine system. However, CDP and CDT turn out to be the two significant process variables in predicting the power of the gas turbine system. Similarly, CDP and GFFR are estimated to be the two significant variables for predicting the turbine heat rate. The significance of CDP, CDT and GFFR is explainable through the operating characteristics of the gas turbine system, which are established through the power generation operation in the industrial environment. CDP and CDT are measured at the outlet of air compressors and establish the thermodynamic conditions of the airflow rate that is injected into the combustion chamber.  GFFR is the amount of gas that is supplied into the combustion chamber and undergoes combustion in the presence of air. The process variables maintain and stabilize temperature of the hot flue gases, which are expanded in the gas turbine to produce power. The work potential of hot flue gases is extracted in a gas turbine, and thermal efficiency and turbine heat rate are measured to account for the energy conversion ratio and thermal energy spent to produce the power from the gas turbine system.

The prediction intervals for the prediction of points made by the three ANN models are computed by inductive conformal prediction \cite{shafer2008tutorial}. The function space built on the observations of the test datasets of the three performance variables is constructed and is labeled the actual test surface in Fig.~\ref{Fig:models}(b). The prediction intervals computed on the three performance variables are utilized to construct upper and lower prediction surfaces with a 95 \% confidence level. We observe that the constructed prediction surfaces enclose the actual test surface, confirming the validity of the upper and lower prediction surfaces to quantify the uncertainty associated with ANN-based predictions for the three performance variables. Moreover, the upper and lower prediction surfaces are in close proximity to the actual test surface, underscoring the excellent modeling performance of the trained ANN models to predict the performance variables of the gas turbine system.

\subsection{Data-centric Optimization of Gas Turbine System}

Industrial gas turbine systems are installed to supply the electrical power in the grid to meet energy demand. In the industrial environment, power supply is prioritized over thermal efficiency and turbine heat rate, since the produced power is sold to the central power purchaser. However, during power generation operations, operators also tend to maximize thermal efficiency and minimize the turbine heat rate, as these performance variables affect the cost of power generation. Therefore, we have formulated an objective function that aims to maximize thermal efficiency and minimize turbine heat, given that the set value of power that is to be produced from the gas turbine system must be satisfied. This mimics the operating scenario of the operation of the gas power plant that is ramped up or ramped down on the available capacity such that the optimal operating conditions to operate the gas turbine system are estimated on the set value of power. However, we have analyzed the formulated objective function and the constraint on power generation under two scenarios: (1) the MAD-OPT framework and (2) when we do not embed the Mahalanobis constraint into the optimization problem. Here, the purpose of analysis of these two types of optimization problem is to demonstrate how embedding and not embedding the domain-specific information into the optimization problems provide "Optimal" solutions that may or may not be implemented in the operation of the industrial systems in general and gas power plants in particular.

\begin{figure}[htp]
    \centering
    \includegraphics[width=1.0\linewidth]{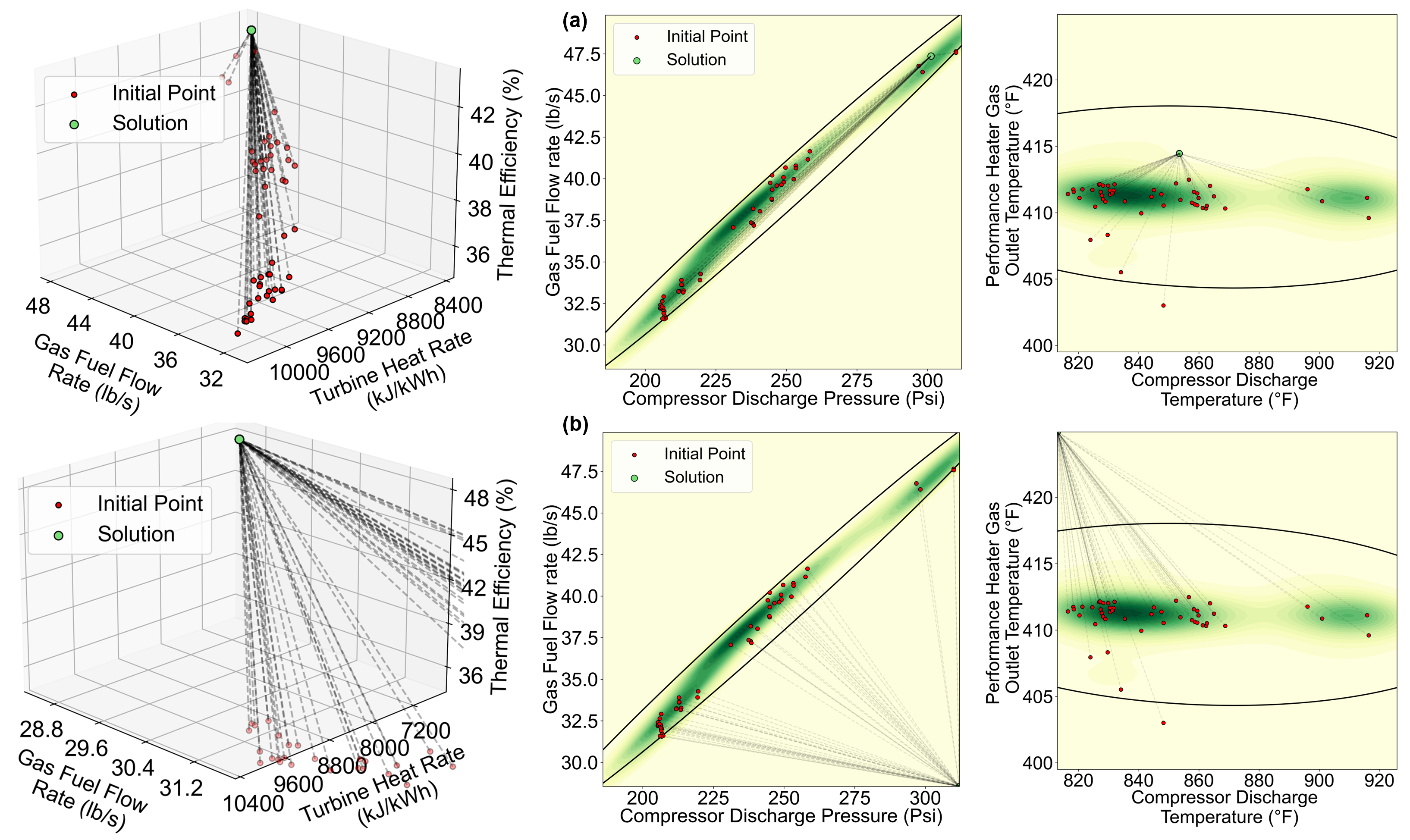}
    \caption{Solving an optimization problem and mapping the feasible optimal solutions for a power set point of 390 MW by (a) the MAD-OPT framework and (b) when the Mahanalobis constraints are not embedded in the optimization problem.}
    \label{Fig:opt_case}
\end{figure}

Fig.~\ref{Fig:opt_case} shows the convergence of the optimization solver to the feasible optimal solution corresponding to the different initial points with respect to the set power value of 390 MW for the MAD-OPT framework (Fig.~\ref{Fig:opt_case}(a)) and when the Mahalanobis constraint is not embedded in the optimization problem (Fig.~\ref{Fig:opt_case}(b). The mapping of the optimal values of the process variables is also depicted against the contour plots which are made on the actual data of the power plant to demonstrate the domain consistency of the optimal solutions under two types of optimization problem. The empiric approach is implemented to estimate the width of the ellipse that captures the data corresponding to the pair of variables: the tighter the ellipse is, the more correlated the pair of variables are, and vice versa. We have taken compressor discharge pressure and gas fuel flow rate as a correlated pair of variables to make a contour plot and to map the optimal solution in the plot. The two correlated variables are selected because they turn out to be the significant variables affecting the power generation from the gas turbine system. Similarly, a weakly correlated pair of variables, compressor discharge temperature \& performance heater gas outlet temperature, is selected to visualize how optimal solutions would be mapped against the contour plot made in the actual data of the gas power plant. 

It is important to mention earlier that we have set the tighter bounds on variables related to ambient conditions near the mean values estimated from the actual data of the power plant for solving the optimization problem in order to determine the optimal values of process variables corresponding to the set value of power of 390 MW. The tighter bounds on the ambient conditions allow simulating an industrial operating scenario corresponding to nearly fixed ambient conditions when capacity discharge of the gas turbine system is going to be changed at a particular instant of time. The tolerance on the Mahalanobis distance is set to 0.9, which serves as a feasible region for the optimization solver to estimate the solution. 

Fig.~\ref{Fig:opt_case}(a) displays the optimal values of gas fuel flow rate (47.4 lb/s), turbine heat rate (8447 kJ / kWh), and thermal efficiency (43.02 \%) corresponding to a set power value of 390 MW, which are estimated by the MAD-OPT framework. The optimal values of the turbine heat rate and thermal efficiency are estimated to be within their operating limits, which were established corresponding to plant's operation. It is also apparent that the estimated optimal solutions for the process variables are mapped within the ellipses, confirming the efficacy of the solutions in terms of their implementation in the operation of the gas turbine system.  

Referring to Fig.~\ref{Fig:opt_case}(b), although the optimization solver achieves a feasible solution on the set power of 390 MW; the optimal values of turbine heat rate (6793 kJ/kWh) and thermal efficiency (47.80 \%) are significantly beyond the operating limits of these two performance variables, as mentioned in Table \ref{table:1}. The solver estimates these optimal values corresponding to the gas fuel flow rate of 28.56 lb/s which is its minimum operating value during the plant's operation. The anomalous optimization results obtained without the Mahalanobis constraint can be identified by visualizing the mappings of optimal solutions related to process conditions in the contour plots made in Fig.~\ref{Fig:opt_case}(b) and these solutions lie outside the operating envelopes. Although we have achieved nearly the same modeling performance on the test datasets for the three variables that demonstrate good generalization performance of the trained ANN models; the integration of parametric ANN models with a "goal-oriented" optimization solver may result in the estimation of a "feasible" optimal solution that may not satisfy the variable correlation structures that were mapped during the training of the ANN models. 

\subsubsection{Stress test—Ramping up the Power Generation from Gas Turbine System}

We also estimate the optimal process conditions when a gas power plant is increased from 185 to 395 MW with a step size of 15 MW. The ramping-up operation is simulated in three sets of ambient conditions, with ambient temperatures of 22 $^{\circ}$C (minimum), 26 $^{\circ}$C (average) and 34 $^{\circ}$C (maximum) taken from the actual data of the gas power plant operation. We also analyze how the integration of the Mahalanobis constraint affects the efficacy of optimal solutions in terms of their domain consistency. Fig.~\ref{Fig:opt_case} compares the solutions profiles estimated during the operation of the increase in power under three different ambient conditions corresponding to (a) the MAD-OPT framework and (b) no integration of the Mahalanobis constraint in the optimization problems.   

Referring to Fig.~\ref{Fig:opt_case}(a), we notice a smooth and increasing trend in compressor discharge pressure and thermal efficiency, while a decreasing trend in turbine heat rate increases when the power set point increases from 185 MW to 395 MW. Thermal efficiency of the gas turbine system is increased from 34.35 \% to 42.99 \% and 34.77 \% to 42.60 \% corresponding to the ambient temperatures of 22 $^{\circ}$C and 34 $^{\circ}$C respectively on ramping up the power. It is important to note here that, overall, thermal efficiency of the gas turbine system is decreased by 0.1 \% when operated at 34 $^{\circ}$C in comparison to the full power ramp-up turbine's operation at 22 $^{\circ}$C. Similarly, gas fuel flow rate is found to be decreased by 0.18 lb/s on operating the gas turbine system at 22 $^{\circ}$C as compared to its operation at 34 $^{\circ}$C during the power ramp-up.

\begin{figure}[htp]
    \centering
    \includegraphics[width=1.1\linewidth]{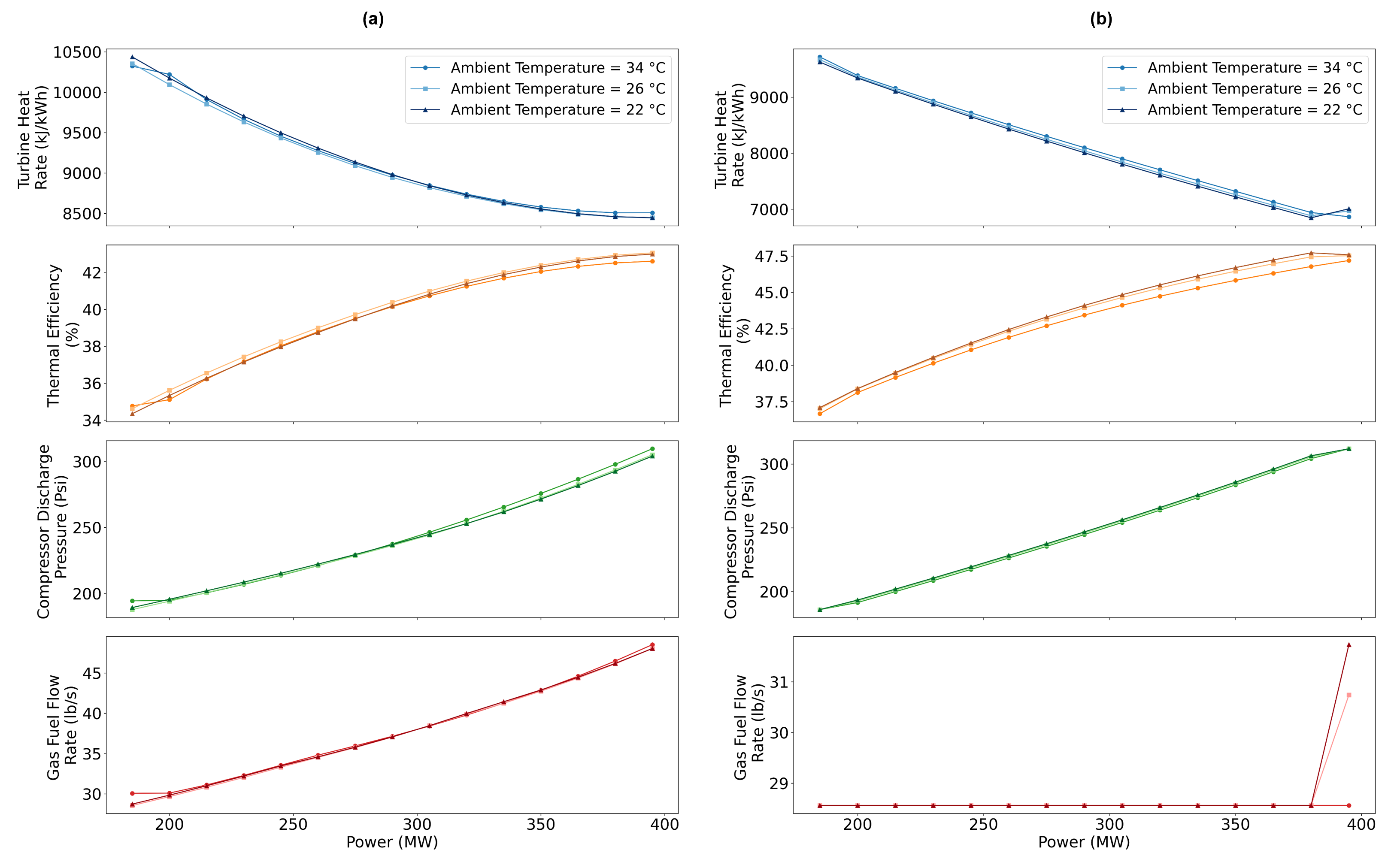}
    \caption{Performance curves obtained on solving the optimization problem for power ramp-up operation of the gas turbine system by (a) the MAD-OPT framework and (b) when the Mahalanobis constraints are not embedded.}
    \label{Fig:per_curves}
\end{figure}

Although we notice smooth and increasing trends for compressor discharge pressure and thermal efficiency, with a decreasing trend for turbine heat rate when power ramp-up is optimized without embedding the Mahalanobis constraint into the optimization problem as shown in Fig.~\ref{Fig:opt_case}(b). The optimization solver consistently estimates 28.56 lb/s as an optimized value of the gas fuel flow rate for almost all set points of power under three ambient conditions. This anomalous value of gas fuel flow rate clearly indicates the domain-inconsistent performance of parametric ANN models when embedded in the optimization problems which lack the integration of domain knowledge. These results reinforce the need to embed domain-informed data-driven constraints into the optimization problem that guide the solver during the solution-search phase for estimating a truly "feasible" and "domain-informed" solution for the optimization problem. Thus, merely relying on the input-output mappings within the ANN model is not sufficient to estimate effective optimal solutions for the optimization problems having ANN models embedded. The quantification of data patterns and the introduction of them as constraints in the optimization problem synergizes with the capabilities of ANN models to accelerate the estimation of domain-informed optimal solutions that may enhance the operational excellence of industrial systems.    

\subsubsection{Robustness Evaluation of the Optimal Solutions}

\begin{figure}[htp]
    \centering
    \includegraphics[width=1.1\linewidth]{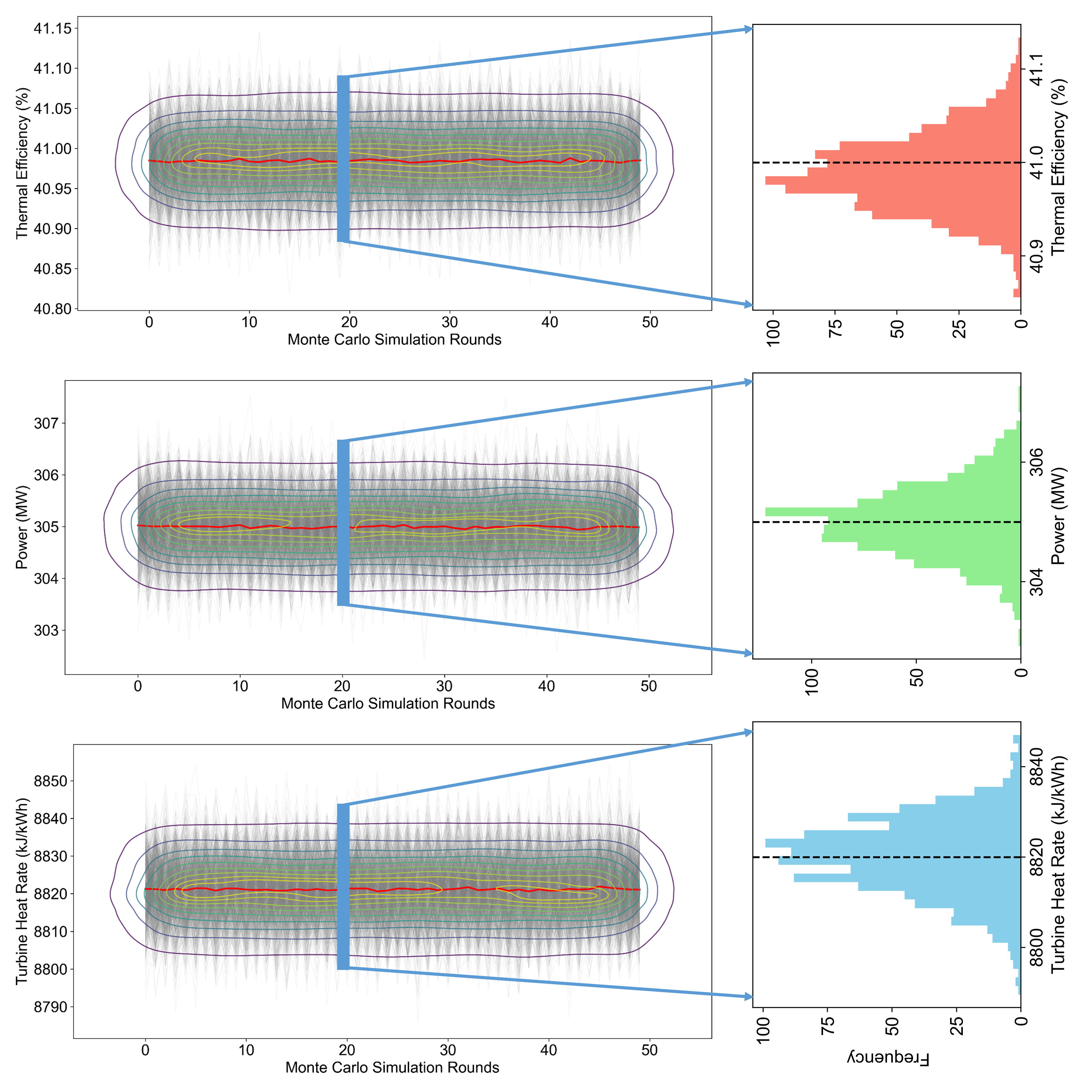}
    \caption{Evaluating the robustness of the optimal solutions obtained through ANN models.}
    \label{Fig:uncer}
\end{figure}

Fig.~\ref{Fig:uncer} shows the variation in the optimal values of thermal efficiency and turbine heat rate corresponding to the set value of power. A mean response is calculated in each simulation round and plotted to visualize its variation across the simulation rounds. A contour plot is also made to visualize the data distribution density for the rounds of simulations. We observe a fairly flat mean response for three performance variables that is quite close to the deterministic solutions, that is, thermal efficiency = 40.98 \%, power = 305 MW and turbine heat rate = 8820 kJ/kWh. The mean response and confidence interval in the highlighted round of Monte simulations are calculated as: thermal efficiency = 40.99 $\pm$ 0.16 \%, power = 304.98 $\pm$ 2.20 MW, turbine heat rate = 8820 $\pm$ 32.81 kJ/kWh. The confidence intervals computed on the perturbed inputs for the three performance variables appear to be tight, demonstrating that the estimated solution is fairly robust to the perturbation in the process input variables.

\subsection{Operating beyond the Designed Capacity}

The ANN models are trained on a subspace of actual data and predict power, thermal efficiency, and turbine heat rate on the test dataset using the following performance metrics: (R$^{2}$ = 1.0, RMSE = 1.50 MW)$_{\text{Power}}$, (R$^{2}$ = 0.95, RMSE = 0.46 \%)$_{\text{Thermal Efficiency}}$ and (R$^{2}$ = 0.96 MW, RMSE = 103 kJ/kWh)$_{\text{Turbine Heat Rate}}$. We observe comparable modeling performance of the trained ANN models on the subspace of actual data as compared to full-space's modeling results reported in the previous section.

 We use the same optimization problem formulation as utilized previously and set the power values of 385 MW, 390 MW, and 395 MW separately for two types of optimization problems: (a) when the Mahalanobis constraint is embedded (MAD-OPT framework) and (b) when the Mahalanobis constraint is not embedded in the optimization problem for extrapolating the neural networks. During extrapolation, our objective is to maximize thermal efficiency and minimize turbine heat rate while estimating the optimal process conditions which satisfy the constraint of a set value of power. We set the upper bounds of the process variables to 1.8 (on a scaled basis) since producing power beyond the design limits may require higher values of the process variables. Moreover, we use the mean values of ambient conditions and slightly perturb them, since these conditions remain nearly constant when power is ramped up or ramped down at the particular instant of time. 
 
\begin{figure}[htp]
    \centering
    \includegraphics[width=1.0\linewidth]{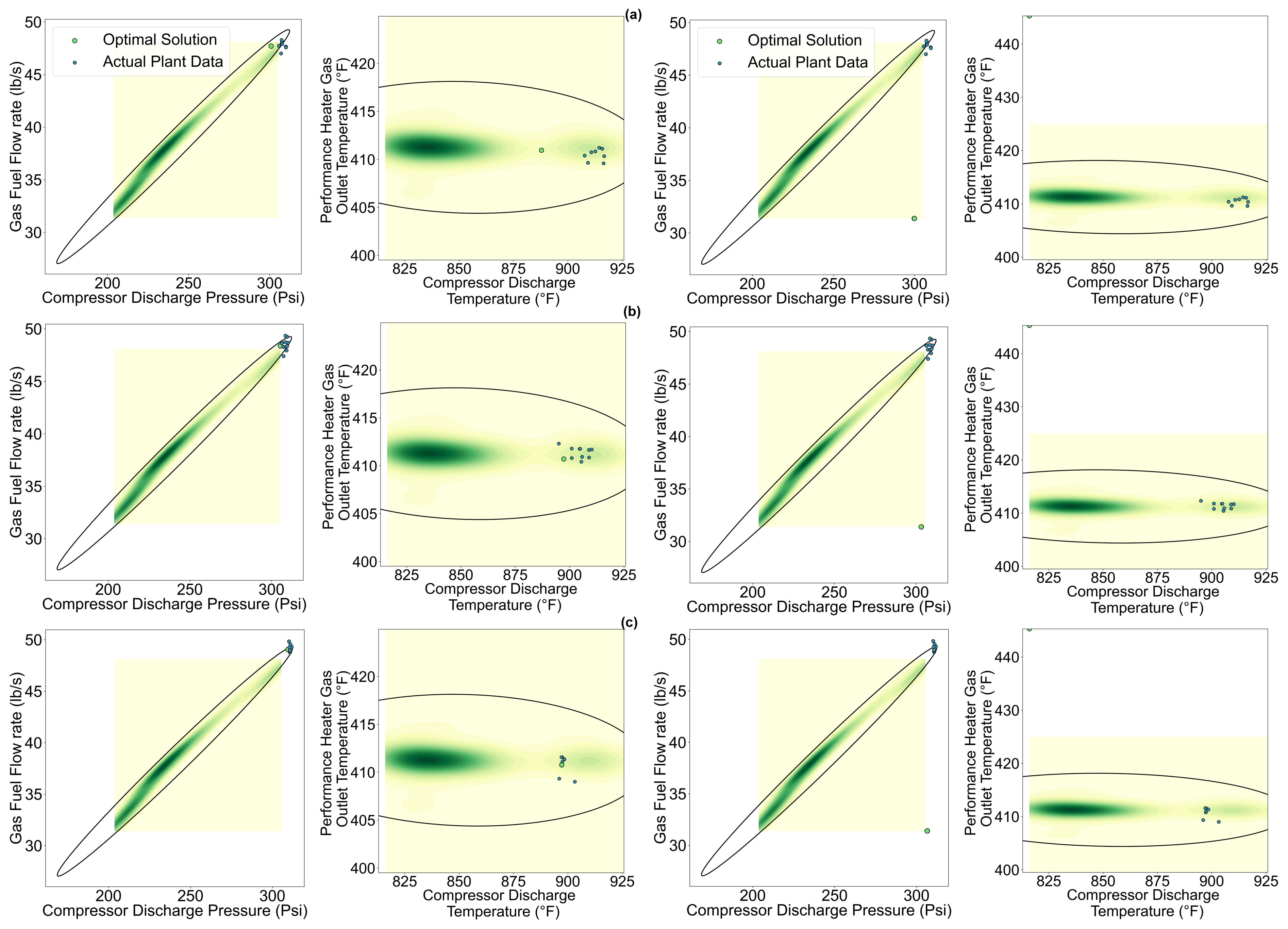}
    \caption{Extrapolating ANN models for producing power beyond the design limit of gas turbine system. The optimization problem is solved with and without the Mahalanobis constraint, and the solutions are mapped for the set values of power of (a) 385 MW, (b) 390 MW, and (c) 395 MW. The pair of plots on the left side corresponds to the extrapolation results obtained by the MAD-OPT framework.} 
    \label{Fig:extr}
\end{figure}

Fig.~\ref{Fig:extr}(a-c) compares the optimization-based extrapolation performance evaluation of the trained ANN models corresponding to the set values of power of 385 MW, 390 MW, and 395 MW, respectively. The optimization problem is solved using the MAD-OPT framework, and the optimal solution mappings are shown on the left side of Fig.~\ref{Fig:extr}. We tune the tolerance parameter related to the Mahalanobis constraint during the extrapolation analysis since it controls the feasible space that the solver can explore during the estimation of solutions. Setting a high tolerance value in combination with large limits on process variables may result in out-of-ellipse mappings of optimal solutions, which will be impractical to implement in the gas turbine system. We fixed the upper bounds of the process variables and tried different values of the tolerance parameter linked to the Mahalanobis constraint such that the set value of the power constraint is satisfied and the optimal solution is mapped within the ellipses of the pair of variables, as shown on the left side of Fig.~\ref{Fig:extr}(a-c). The optimal value of the tolerance parameter is estimated to be 0.4, 0.45 and 0.6, corresponding to set values of power of 385 MW, 390 MW and 395 MW respectively. These values are estimated based on the in-ellipse mappings of optimal solutions, as well as satisfaction of the set-power constraint that reflects confidence in extrapolating the neural network with the domain information and variable dependencies captured through the Mahalanobis constraint. 

We also compare the optimal values of the process conditions with the corresponding to actual values of process conditions that we had already separated during sub-space sampling from the actual data of the gas turbine system. The comparison reveals that compressor discharge pressure and gas fuel flow rate need to be set at higher values while producing power above the design limit (380 MW). The yellow-shaded rectangle is the subspace of the two process conditions on which ANN models have been trained. The optimal solutions corresponding to the set power values of 385 MW, 390 MW, and 395 MW lie not only outside the subspace, but also close to the actual data of the power plant, which serve as ground truth to verify the accuracy of the extrapolation analyzes. However, the optimal solutions estimated for compressor discharge temperature and performance heater gas outlet temperature remain in-ellipse under three set values of power and are not extrapolated beyond the subspace. This can be attributed to the low significance of these process variables in ANN-based predictions and the weak correlation between them, which makes them set at operating levels such that their correlation structure is satisfied.

We also estimate the optimal process conditions for extrapolating the power generation from the gas turbine system such that the Mahalanobis constraint is not embedded in the optimization framework. Under such conditions, the mappings of the optimal solutions are made and are shown on the right side of Fig.~\ref{Fig:extr}(a-c) corresponding to the set values of power of 385 MW, 390 MW and 395 MW, respectively. The optimal solutions for correlated pairs of variables are mapped outside the ellipse, and interestingly, the solver does not enable the ANN to extrapolate well even though the upper bounds on the process variables were set to 1.8, the same as the Mahalanobis constraint-based extrapolation analysis. Similarly, the optimal values of the weakly correlated pair of process variables are mapped significantly outside their subspace region and seem not to follow the structure of the variable dependencies. 

\subsubsection{Robustness Evaluation of ANN-based Extrapolated Solutions}

We apply the same methodology for the robustness evaluation of optimal solutions using Monte Carlo simulations. The only difference is the noise level, which is taken as 1.2 \% of the standard deviation of the actual data associated with the process variables of the gas turbine system. A relatively larger noise level can cover a wide space around the optimal solution and evaluate the sensitivity of the optimal solutions to the perturbations in the process conditions. The confidence intervals are calculated for the robustness evaluation of extrapolated solutions.

\begin{figure}[htp]
    \centering
    \includegraphics[width=0.6\linewidth]{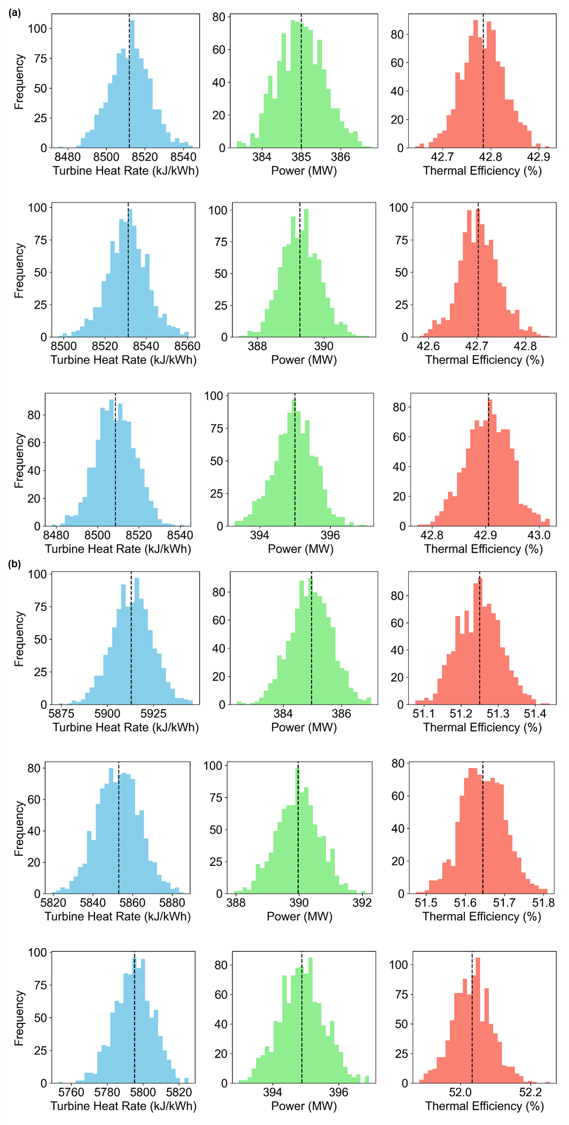}
    \caption{Quantifying the robustness of extrapolated solutions corresponding to set values of power of 385 MW, 390 MW and 395 MW by (a) the MAD-OPT framework and (b) when the Mahalanobis constraint is not embedded in an optimization problem.}
    \label{Fig:extr_uncer}
\end{figure}

Fig.~\ref{Fig:extr_uncer} compares the Monte Carlo simulations-based distributions of responses around the deterministic solution estimate by (a) the MAD-OPT framework and (b) the Mahalanobis constraint is not embedded in the optimization problem. The confidence intervals are also calculated for the extrapolated set values of power of 385 MW, 390 MW, and 395 MW. Referring to Fig.~\ref{Fig:extr_uncer}(a), mean turbine heat rate is changed from 8512 $\pm$ 40 kJ/kWh to 8508 $\pm$ 36 kJ / kWh, the power is varied from 385 $\pm$ 2.2 MW to 395 $\pm$ 2.2 MW and the thermal efficiency is increased from 42.78 $\pm$ 0.17 \% to  42.90 $\pm$ 0.16 \%. The confidence intervals are fairly tight, and the estimates are based on the perturbed process variables. We also compute the confidence intervals on the optimal solutions when estimated without the Mahalanobis constraint. Although the optimal values of thermal efficiency and turbine heat rate are significantly away from their operating ranges, we observe relatively higher confidence intervals on their estimates. We computed the confidence interval of 42 kJ / kWh, 42 kJ / kWh and 41 kJ / kWh for the turbine heat rate, 2.8 MW, 2.6 MW and 2.7 MW for power, and 0.22 \%, 0.22 \% and 0.23 \% for thermal efficiency, corresponding to set power values of 385 MW, 390 MW and 395 MW, respectively. The relatively higher confidence intervals computed for extrapolated solutions without the Mahalanobis constraint may be attributed to solutions convergence significantly away from the trained space of ANN models, and models may exhibit higher sensitive responses, leading to relatively larger confidence intervals than those of solutions estimated closer to the trained space of ANN models.    

The systematic comparison of optimization-based extrapolation analysis carried out with and without the integration of domain-specific information reveals that ANN models can be made to extrapolate with reasonable accuracy and exhibit lower predictive uncertainty, which stems from following the variable dependencies that exist in the data. The challenge remains to identify data-driven constraints that effectively capture domain knowledge. However, domain-knowledge constrained optimization problems can be a way forward towards AI adoption for the operation excellence of the gas turbine system.

\section{Conclusion}

This study demonstrates the successful development and implementation of the MAD-OPT framework for optimizing gas turbine operations, addressing the critical challenge of the adoption of artificial intelligence (AI) in safety-critical industrial systems. The effectiveness of the framework is evidenced by the ANN models that achieve a prediction accuracy of more than 95\% for power, thermal efficiency and turbine heat rate, with reliability confirmed by uncertainty analysis based on inductive conformal prediction. SHAP analysis identified compressor discharge pressure, compressor discharge temperature, and gas fuel flow rate as primary operational determinants. Under varying ambient conditions, the framework revealed that the increase in power from 185 MW to 395 MW yielded a 0.1 percentage point increase in thermal efficiency at 22°C compared to 34°C, with robust optimal values of 40.99 ± 0.16\% for thermal efficiency and 8820 ± 32.81 kJ/kWh for turbine heat rate at 305 MW and 26 °C. In particular, the MAD-OPT framework successfully extrapolated beyond the 380 MW limit of the training data to accurately estimate process conditions at 385-395 MW, while maintaining domain consistency through the Mahalanobis constraint, a feature absent from conventional optimization approaches. These results establish MAD-OPT as a promising framework for integrating data-driven domain knowledge with statistical constraints, potentially accelerating the responsible adoption of AI in safety-critical industrial applications.

\section*{Limitations and future work}

The MAD-OPT framework is built on the Mahalanobis distance-based constraint and the ANN models, which is the main limitation of this research. In the future, innovative and purely data-driven constraints can be developed that quantify the nonlinear dynamics of the system from the data and are embedded in optimization problem. Moreover, the effect of different AI models on the quality of solutions under the same constraints can also be investigated.

\section*{Acknowledgement}
Waqar Muhammad Ashraf acknowledges that this work was supported by The Alan Turing Institute’s Enrichment Scheme/Turing Studentship Scheme. Waqar Muhammad Ashraf also acknowledges the funding (CMMS-PhD-2021-006) received from The Punjab Education Endowment Fund (PEEF) for his PhD at University College London, UK. The authors would like to extend their appreciation to the Deanship of Scientific Research at King Saud University for funding this work through ISPP Program (ISPP25-3).

\section*{Author Contributions}
\textbf{Waqar Muhammad Ashraf:} Conceptualization, Methodology, Data Curation, Software, Validation, Formal Analysis, Writing – Original Draft, \textbf{Amir H. Keshavarzzadeh:}  Validation, Writing – Original Draft, \textbf{Abdulelah S. Alshehri:} Conceptualization, Methodology, Funding Acquisition, Writing – Review \& Editing, \textbf{Abdulrahman bin Jumah:} Conceptualization, Methodology, Funding Acquisition, Writing – Review \& Editing, \textbf{Vivek Dua:} Conceptualization, Writing – Review \& Editing, Project Administration, \textbf{Ramit Debnath:} Validation, Writing – Review \& Editing

\section*{Data availability statement}
The code written for this research is available at: https://github.com/Waqar9871/MAD-OPT.git

\section*{Declaration of Competing Interests}
The authors declare no competing interests.

\bibliographystyle{unsrt}  
\bibliography{references}  

\begin{thebibliography}{10}

\bibitem{davis2018net}
Steven~J Davis, Nathan~S Lewis, Matthew Shaner, Sonia Aggarwal, Doug Arent, In{\^e}s~L Azevedo, Sally~M Benson, Thomas Bradley, Jack Brouwer, Yet-Ming Chiang, et~al.
\newblock Net-zero emissions energy systems.
\newblock {\em Science}, 360(6396):eaas9793, 2018.

\bibitem{bistline2022role}
John~ET Bistline and David~T Young.
\newblock The role of natural gas in reaching net-zero emissions in the electric sector.
\newblock {\em Nature Communications}, 13(1):4743, 2022.

\bibitem{wilson2018rapid}
IA~Grant Wilson and Iain Staffell.
\newblock Rapid fuel switching from coal to natural gas through effective carbon pricing.
\newblock {\em Nature Energy}, 3(5):365--372, 2018.

\bibitem{ieaai}
IEA.
\newblock Energy and ai.
\newblock {\em IEA, Paris https://www.iea.org/reports/energy-and-ai, Licence: CC BY 4.0}, 2025.

\bibitem{deb2022machine}
Smita Deb, Sahil Sidheekh, Christopher~F Clements, Narayanan~C Krishnan, and Partha~S Dutta.
\newblock Machine learning methods trained on simple models can predict critical transitions in complex natural systems.
\newblock {\em Royal Society Open Science}, 9(2):211475, 2022.

\bibitem{vlachas2022multiscale}
Pantelis~R Vlachas, Georgios Arampatzis, Caroline Uhler, and Petros Koumoutsakos.
\newblock Multiscale simulations of complex systems by learning their effective dynamics.
\newblock {\em Nature Machine Intelligence}, 4(4):359--366, 2022.

\bibitem{qi2020using}
Di~Qi and Andrew~J Majda.
\newblock Using machine learning to predict extreme events in complex systems.
\newblock {\em Proceedings of the National Academy of Sciences}, 117(1):52--59, 2020.

\bibitem{dos2024co}
Leandro dos Santos~Coelho, Helon Vicente~Hultmann Ayala, and Viviana~Cocco Mariani.
\newblock Co and nox emissions prediction in gas turbine using a novel modeling pipeline based on the combination of deep forest regressor and feature engineering.
\newblock {\em Fuel}, 355:129366, 2024.

\bibitem{saeed2023electrical}
Mohammed~A Saeed, El-Sayed~M El-Kenawy, Abdelhameed Ibrahim, Abdelaziz~A Abdelhamid, Marwa~M Eid, Faten~Khalid Karim, Doaa~Sami Khafaga, and Laith Abualigah.
\newblock Electrical power output prediction of combined cycle power plants using a recurrent neural network optimized by waterwheel plant algorithm.
\newblock {\em Frontiers in Energy Research}, 11:1234624, 2023.

\bibitem{siddiqui2021power}
Raheel Siddiqui, Hafeez Anwar, Farman Ullah, Rehmat Ullah, Muhammad~Abdul Rehman, Naveed Jan, and Fawad Zaman.
\newblock Power prediction of combined cycle power plant (ccpp) using machine learning algorithm-based paradigm.
\newblock {\em Wireless Communications and Mobile Computing}, 2021(1):9966395, 2021.

\bibitem{gas_turbine_co_and_nox_emission_data_set_551}
Pnar Tfekci and Heysem Kaya.
\newblock {Gas Turbine CO and NOx Emission Data Set}.
\newblock UCI Machine Learning Repository, 2019.
\newblock {DOI}: https://doi.org/10.24432/C5WC95.

\bibitem{combined_cycle_power_plant_294}
Pnar Tfekci and Heysem Kaya.
\newblock {Combined Cycle Power Plant}.
\newblock UCI Machine Learning Repository, 2014.
\newblock {DOI}: https://doi.org/10.24432/C5002N.

\bibitem{ashraf2025domain}
Waqar~Muhammad Ashraf, Vivek Dua, and Ramit Debnath.
\newblock Domain consistent industrial decarbonisation of global coal power plants.
\newblock {\em arXiv preprint arXiv:2503.03571}, 2025.

\bibitem{shobeiry2024ai}
Seyed~Mohammad Shobeiry.
\newblock Ai-enabled modern power systems: Challenges, solutions, and recommendations.
\newblock In {\em Artificial Intelligence in the Operation and Control of Digitalized Power Systems}, pages 19--67. Springer, 2024.

\bibitem{osegi2023evolutionary}
Emmanuel~N Osegi, Zaid~OO Jagun, Cornelius~C Chujor, Vincent~IE Anireh, Biobele~A Wokoma, and Otonye Ojuka.
\newblock An evolutionary programming technique for evaluating the effect of ambient conditions on the power output of open cycle gas turbine plants-a case study of afam gt13e2 gas turbine.
\newblock {\em Applied Energy}, 349:121661, 2023.

\bibitem{wu2023prediction}
Wei Wu, Yan-Ting Lin, Po-Hsuan Liao, Muhammad Aziz, and Po-Chih Kuo.
\newblock Prediction of co--nox emissions from a natural gas power plant using proper machine learning models.
\newblock {\em Energy Technology}, 11(7):2300041, 2023.

\bibitem{manatura2025gas}
Kanit Manatura, Nawaporn Rummith, Benjapon Chalermsinsuwan, Namfon Samsalee, Wei-Hsin Chen, Kankamon Phookronghin, and Sutthipoj Wongrerkdee.
\newblock Gas turbine heat rate prediction in combined cycle power plant using artificial neural network.
\newblock {\em Thermal Science and Engineering Progress}, page 103301, 2025.

\bibitem{designs7040085}
Marwan Al-Shami, Omar Mohamed, and Wejdan Abu~Elhaija.
\newblock Energy-efficient control of a gas turbine power generation system.
\newblock {\em Designs}, 7(4), 2023.

\bibitem{mamlook2011fuzzy}
Rustom Mamlook, Omar Badran, Abdullah Aljumah, Abdulaziz~S Almazyad, Taisir Eldos, and Emad Abdulhadi.
\newblock Fuzzy logic controller to improve parameters affecting gas turbines power generation.
\newblock {\em Clean Technologies and Environmental Policy}, 13:821--829, 2011.

\bibitem{bertini2010fuzzy}
Ilaria Bertini, Alessandro Pannicelli, and Stefano Pizzuti.
\newblock Fuzzy optimization of start-up operations for combined cycle power plants.
\newblock In {\em Soft Computing Models in Industrial and Environmental Applications, 5th International Workshop (SOCO 2010)}, pages 153--160. Springer, 2010.

\bibitem{mohamed2016predictive}
Omar Mohamed, Jihong Wang, Ashraf Khalil, and Marwan Limhabrash.
\newblock Predictive control strategy of a gas turbine for improvement of combined cycle power plant dynamic performance and efficiency.
\newblock {\em SpringerPlus}, 5:1--20, 2016.

\bibitem{hou2018novel}
Guolian Hou, Linjuan Gong, Xiaoyan Dai, Mengyi Wang, and Congzhi Huang.
\newblock A novel fuzzy model predictive control of a gas turbine in the combined cycle unit.
\newblock {\em Complexity}, 2018(1):6468517, 2018.

\bibitem{hou2020fuzzy}
Guolian Hou, Linjuan Gong, Congzhi Huang, and Jianhua Zhang.
\newblock Fuzzy modeling and fast model predictive control of gas turbine system.
\newblock {\em Energy}, 200:117465, 2020.

\bibitem{ratz2024identifying}
Martin R{\"a}tz, Patrick Henkel, Phillip Stoffel, Rita Streblow, and Dirk M{\"u}ller.
\newblock Identifying the validity domain of machine learning models in building energy systems.
\newblock {\em Energy and AI}, 15:100324, 2024.

\bibitem{braiek2025machine}
Houssem~Ben Braiek and Foutse Khomh.
\newblock Machine learning robustness: A primer.
\newblock In {\em Trustworthy AI in Medical Imaging}, pages 37--71. Elsevier, 2025.

\bibitem{freiesleben2023beyond}
Timo Freiesleben and Thomas Grote.
\newblock Beyond generalization: a theory of robustness in machine learning.
\newblock {\em Synthese}, 202(4):109, 2023.

\bibitem{goyal2020prediction}
Vipul Goyal, Mengyu Xu, Jayanta Kapat, and Ladislav Vesely.
\newblock Prediction of gas turbine performance using machine learning methods.
\newblock In {\em Turbo Expo: Power for Land, Sea, and Air}, volume 84157, page V006T09A004. American Society of Mechanical Engineers, 2020.

\bibitem{saleel2021forecasting}
C~Ahamed Saleel.
\newblock Forecasting the energy output from a combined cycle thermal power plant using deep learning models.
\newblock {\em Case Studies in Thermal Engineering}, 28:101693, 2021.

\bibitem{tufekci2014prediction}
P{\i}nar T{\"u}fekci.
\newblock Prediction of full load electrical power output of a base load operated combined cycle power plant using machine learning methods.
\newblock {\em International Journal of Electrical Power \& Energy Systems}, 60:126--140, 2014.

\bibitem{liu2020gas}
Zuming Liu and Iftekhar~A Karimi.
\newblock Gas turbine performance prediction via machine learning.
\newblock {\em Energy}, 192:116627, 2020.

\bibitem{ashraf2024driving}
Waqar~Muhammad Ashraf and Vivek Dua.
\newblock Driving towards net-zero from the energy sector: Leveraging machine intelligence for robust optimization of coal and combined cycle gas power stations.
\newblock {\em Energy Conversion and Management}, 314:118645, 2024.

\bibitem{csaji2001approximation}
Bal{\'a}zs~Csan{\'a}d Cs{\'a}ji et~al.
\newblock Approximation with artificial neural networks.
\newblock {\em Faculty of Sciences, Etvs Lornd University, Hungary}, 24(48):7, 2001.

\bibitem{goswami2024fault}
Umang Goswami, Hariprasad Kodamana, and Manojkumar Ramteke.
\newblock Fault detection using graph neural differential auto-encoders (gndae).
\newblock {\em Computers \& Chemical Engineering}, 189:108775, 2024.

\bibitem{kumar2025learning}
Deepak Kumar, Vinayak Dixit, Manojkumar Ramteke, and Hariprasad Kodamana.
\newblock Learning system physics using symbolic neural integration (synism) with applications to chemical processes.
\newblock {\em Industrial \& Engineering Chemistry Research}, 2025.

\bibitem{ansar2025comparison}
Talha Ansar and Waqar~Muhammad Ashraf.
\newblock Comparison of kolmogorov--arnold networks and multi-layer perceptron for modelling and optimisation analysis of energy systems.
\newblock {\em Energy and AI}, 20:100473, 2025.

\bibitem{kumar2023variance}
Deepak Kumar, Umang Goswami, Hariprasad Kodamana, Manojkumar Ramteke, and Prakash~Kumar Tamboli.
\newblock Variance-capturing forward-forward autoencoder (vffae): A forward learning neural network for fault detection and isolation of process data.
\newblock {\em Process Safety and Environmental Protection}, 178:176--194, 2023.

\bibitem{rumelhart1986learning}
David~E Rumelhart, Geoffrey~E Hinton, and Ronald~J Williams.
\newblock Learning representations by back-propagating errors.
\newblock {\em nature}, 323(6088):533--536, 1986.

\bibitem{gueddar2012novel}
Taoufiq Gueddar and Vivek Dua.
\newblock Novel model reduction techniques for refinery-wide energy optimisation.
\newblock {\em Applied energy}, 89(1):117--126, 2012.

\bibitem{jamil2024machine}
Muhammad~Ahmad Jamil, Waqar~Muhammad Ashraf, Nida Imtiaz, Ben~Bin Xu, Syed~M Zubair, Haseeb Yaqoob, Muhammad Imran, and Muhammad~Wakil Shahzad.
\newblock Machine learning-based process design of a novel sustainable cooling system.
\newblock {\em Energy Conversion and Management}, 319:118941, 2024.

\bibitem{haykin2009neural}
Simon Haykin.
\newblock {\em Neural networks and learning machines, 3/E}.
\newblock Pearson Education India, 2009.

\bibitem{kingma2014adam}
Diederik~P Kingma and Jimmy Ba.
\newblock Adam: A method for stochastic optimization.
\newblock {\em arXiv preprint arXiv:1412.6980}, 2014.

\bibitem{muhammad2020optimization}
Waqar Muhammad~Ashraf, Ghulam Moeen~Uddin, Ahmad Hassan~Kamal, Muhammad Haider~Khan, Awais~Ahmad Khan, Hassan Afroze~Ahmad, Fahad Ahmed, Noman Hafeez, Rana Muhammad Zawar~Sami, Syed Muhammad~Arafat, et~al.
\newblock Optimization of a 660 mwe supercritical power plant performance—a case of industry 4.0 in the data-driven operational management. part 2. power generation.
\newblock {\em Energies}, 13(21):5619, 2020.

\bibitem{lundberg2017unified}
Scott~M Lundberg and Su-In Lee.
\newblock A unified approach to interpreting model predictions.
\newblock {\em Advances in neural information processing systems}, 30, 2017.

\bibitem{tariq2025explainable}
Rasikh Tariq, Juan~A Recio-Garcia, Armando~J Cetina-Qui{\~n}ones, Mauricio~G Orozco-del Castillo, and Ali Bassam.
\newblock Explainable artificial intelligence twin for metaheuristic optimization: double-skin facade with energy storage in buildings.
\newblock {\em Journal of Computational Design and Engineering}, page qwaf015, 2025.

\bibitem{mclachlan1999mahalanobis}
Goeffrey~J McLachlan.
\newblock Mahalanobis distance.
\newblock {\em Resonance}, 4(6):20--26, 1999.

\bibitem{ghorbani2019mahalanobis}
Hamid Ghorbani.
\newblock Mahalanobis distance and its application for detecting multivariate outliers.
\newblock {\em Facta Universitatis, Series: Mathematics and Informatics}, pages 583--595, 2019.

\bibitem{gallego2013mahalanobis}
Guillermo Gallego, Carlos Cuevas, Raul Mohedano, and Narciso Garcia.
\newblock On the mahalanobis distance classification criterion for multidimensional normal distributions.
\newblock {\em IEEE Transactions on Signal Processing}, 61(17):4387--4396, 2013.

\bibitem{gill2005snopt}
Philip~E Gill, Walter Murray, and Michael~A Saunders.
\newblock Snopt: An sqp algorithm for large-scale constrained optimization.
\newblock {\em SIAM review}, 47(1):99--131, 2005.

\bibitem{shafer2008tutorial}
Glenn Shafer and Vladimir Vovk.
\newblock A tutorial on conformal prediction.
\newblock {\em Journal of Machine Learning Research}, 9(3), 2008.

\end{thebibliography}



\end{document}